\RequirePackage{fix-cm}
\documentclass[twocolumn]{svjour3}          
\smartqed  
\usepackage{graphicx}
\usepackage{cite}  
\usepackage[numbers,sort&compress]{natbib}
\usepackage{amsmath,amssymb,amsfonts}
\usepackage{textcomp}
\usepackage{xcolor}    
\usepackage{graphicx}  
\usepackage{multirow}  
\usepackage{url}
\usepackage{booktabs}
\usepackage[misc]{ifsym}
\begin{document}

\title{Towards End-to-end Car License Plate Location and Recognition in Unconstrained Scenarios
}


\author{Shuxin Qin  \textsuperscript{1,~\Letter} \and
        Sijiang Liu \textsuperscript{2}
}


\institute{
	\begin{itemize}
		\item[\textsuperscript{\Letter}] {Shuxin Qin} \\
		\email{qinshuxin@ainnovation.com}
		\at
		\item[\textsuperscript{1}] the Department of Research and Development, AInnovation Technology Co., Ltd., No.6 Xingzhi Road, Qixia District, Nanjing, China.
		\item[\textsuperscript{2}] Nanjing University of Posts and Telecommunications, No.9 Wenyuan Road, Qixia District, Nanjing, China.
	\end{itemize}
}	
\date{Received: date / Accepted: date}

\maketitle

\begin{abstract}
Benefiting from the rapid development of convolutional neural networks, the performance of car license plate detection and recognition has been largely improved. Nonetheless, most existing methods solve detection and recognition problems separately, and focus on specific scenarios, which hinders the deployment for real-world applications. To overcome these challenges, we present an efficient and accurate framework to solve the license plate detection and recognition tasks simultaneously. It is a lightweight and unified deep neural network, that can be optimized end-to-end and work in real-time. Specifically, for unconstrained scenarios, an anchor-free method is adopted to efficiently detect the bounding box and four corners of a license plate, which are used to extract and rectify the target region features. Then, a novel convolutional neural network branch is designed to further extract features of characters without segmentation. Finally, the recognition task is treated as sequence labeling problems, which are solved by Connectionist Temporal Classification (CTC) directly. Several public datasets including images collected from different scenarios under various conditions are chosen for evaluation. Experimental results indicate that the proposed method significantly outperforms the previous state-of-the-art methods in both speed and precision.
\end{abstract}

\section{Introduction}
\label{intro}
In modern life, automatic license plate recognition (ALPR) as a very widely used technology, plays an important role in intelligent transportation system (ITS). It can be used in parking management, security surveillance, traffic control and other fields \cite{Du2013}. A conventional ALPR system usually consists of vehicle detection, license plate location and recognition \cite{Silva_2018_ECCV, BJORKLUND2019134, 1909.01754, 9204696, SILVA2020102773, 9092977}. Vehicle detection is an optional stage that can be considered as the pre-processing of license plate location. Then, object detection methods can be adopted to obtain the bounding box of each license plate. The recognition stage usually reads the segmented license plate area and decodes each character by using segmentation-based or segmentation-free methods. Accordingly, two key sub-tasks (license plate detection and license plate recognition) should be solved by the ALPR systems. With the rapid development of convolutional neural networks, much work has been done to solve at least one of the twos.

Although ALPR has been widely used and achieved good results under specific situations, there are still various challenges for unconstrained scenarios. For example, due to the influence of environmental factors such as illumination and weather, poor quality images of the license plate lead to a decrease in the accuracy of recognition. In addition, the large tilt degree of the license plate shooting leads to inaccurate segmentation for both license plate and characters, which directly lower the accuracy of recognition. Moreover, running speed has always been an important measure for the ALPR systems, especially for edge devices and smart camera applications. That means high accuracy achieved by using larger networks or more complex processing will also lead to a decrease in speed. 

Most of previous works focus on one of the two sub-tasks or solve them separately \cite{Yuan2017, Xie2018, Silva_2018_ECCV, Zhuang_2018_ECCV, Liu2019, Shemarry2020}. For license plate location sub-task, general object detection methods \cite{yolov3, Liu_2016_ECCV, Ren2017} are usually adopted to generate the bounding boxes. However, a single bounding box could not locate the accurate areas when there is a large tilt or rotate degree of the license plate shooting. For recognition sub-task, segmentation-based methods depends heavily on the quality of character segmentation which is sensitive to unconstrained conditions. On the contrary, segmentation-free methods usually label the license plate characters directly by using Recurrent Neural Network (RNN) methods \cite{WANG2018149,Li2019} which is relatively time-consuming.

Only a few of previous works try to solve the two sub-tasks by a single network\cite{Li2019, Xu_2018_ECCV}. These methods usually use shared convolutional features and multi-task learning strategy for efficiency. Compared with the two-step (detection, recognition) solutions, the end-to-end training manner is more concise and efficient when proper learning strategies are applied. In addition, comparable performance can be achieved with similar processing modules (such as backbone, RNN). 

Generally, we argue that there are still three main challenges that hinder the deployment of the ALPR systems to real-world applications. Specifically, the first challenge is the adaptability and generalization ability for unconstrained scenarios. The second one is the trade-off between speed and accuracy. The convenience of model training and deployment is the third one.

\begin{figure}[!t]
	\centering
	\includegraphics[width=0.48\textwidth]{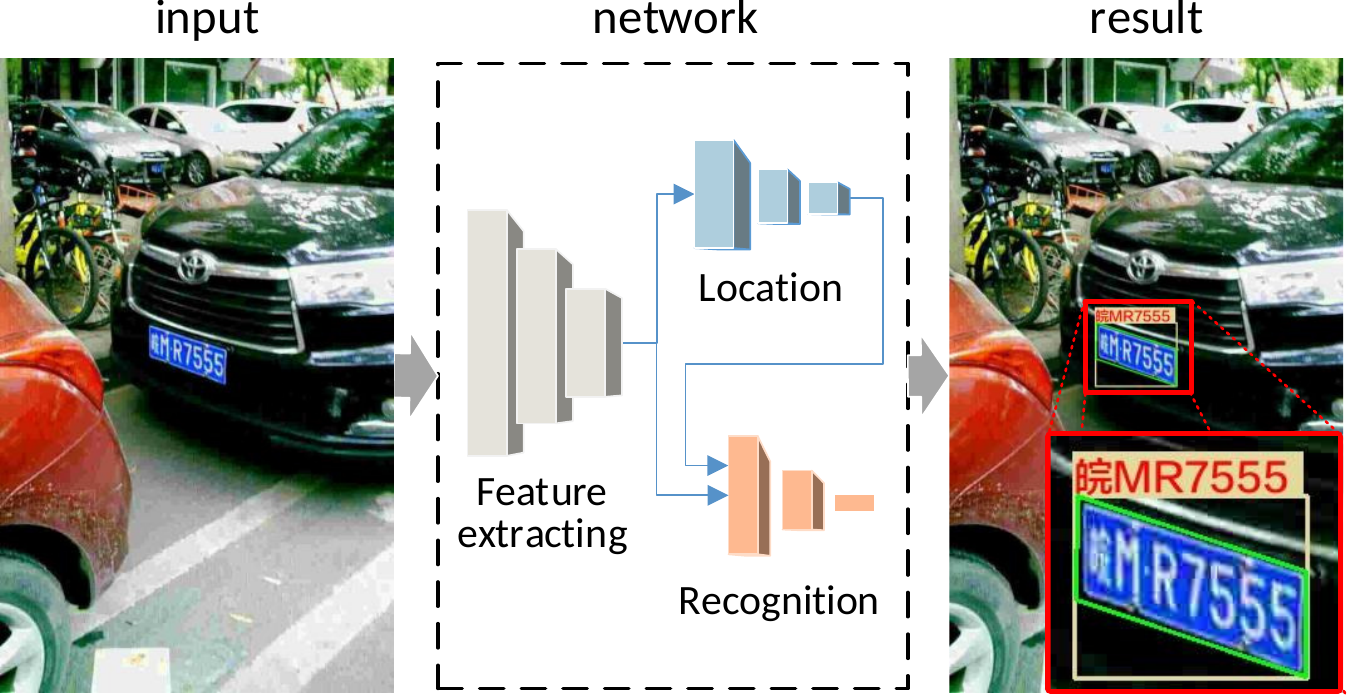}
	\caption{Schematic overview of our proposed framework. The input image is fed to a single neural network that consists of feature extracting, location and recognition. The final result includes three parts: a bounding box, four corners, and the characters.}
	\label{fig1}
\end{figure}

To overcome these challenges, we propose an efficient and accurate framework to solve the two sub-tasks simultaneously. We choose a lightweight deep neural network as the backbone to efficiently extract shared features that are used for both license plate location and recognition. For unconstrained scenarios, both bounding box and four corners of the license plate are regressed by the detection sub-task. The license plate area located by the corners is not sensitive to shooting angle and more accurate than that located by bounding box. Then, rectified features are calculated for license plate recognition which is treated as sequence labeling problems. The recognition sub-task is also based on a very small network taking rectified features as inputs. The whole network can be trained end-to-end and work in real-time. The end-to-end manner and sequence-level annotations make the training and deployment more convenient compared with two-stage methods and segmentation-based methods. Compared with [3], the differences are as follows: 1) Pipeline. We provide an end-to-end network that can read license plate in a single forward pass, while their method is a cascade architecture that includes two relatively independent networks. 2) Feature extraction. We use typical FPN for feature extracting and fusing. The input is a single image. The multi-scale features are extracted and fused for location and recognition. Differently, their scale-pyramid method processes each scaled image separately. The inputs are multi-scale images. They fused the detection results of these scales to get the final result. 3) Location. Our location network is based on an anchor-free method using fully convolutional layers, while their method is based on fully connected layers. 4) Recognition. We treat plate recognition as sequence labeling, while they treat it as character detection and classification problems. Fig. \ref{fig1} describes a schematic overview of our proposed framework.

In summary, the main contributions of this paper are as follows: 1) We present a novel end-to-end trainable network to solve license plate location and recognition tasks simultaneously by using shared features and multi-task learning strategy. It has a much simpler yet effective pipeline compared with the two-step solutions. 2) For unconstrained scenarios, an anchor-free method is designed to efficiently predict both bounding box and four corners of a license plate. The accurate location and rectification can significantly improve the recognition performance. 3) The proposed method is an efficient and accurate network that can work in real-time. A lightweight backbone is used for efficiency. For license plate location, non-maximum suppression (NMS) is removed by using anchor-free method, which can also reduce the processing time significantly. 4) We also introduce a small CNN branch for license plate recognition without segmentation. It is more efficient compared with RNN-based methods.

The rest of this paper is structured as follows. Section \ref{work} provides a brief discussion on related work. Section \ref{method} describes the details of proposed method. Section \ref{experiment} reports the experimental results. Finally, conclusions are drawn in Section \ref{conclusion}.

\section{Related work}
\label{work}
As mentioned above, the ALPR systems includes two key sub-tasks: license plate location and recognition. The previous works can be divided into three broad categories: license plate location, license plate recognition and end-to-end methods.

\subsection{License plate location}

License plate location is to find the position of plate in a digital image. Generally, bounding box is used to represent a location. Before deep learning technology was widely used, traditional methods for location are typically based on edge, color, texture, or character\cite{Li2019}. In \cite{Du2013}, traditional methods are reviewed and compared from different perspectives. Benefiting from the development of deep learning technology and computing power in recent years, object detection methods based on deep neural networks are widely adopted. In \cite{Zhao2019}, a review of deep learning-based object detection frameworks are provided. 

\subsubsection{Traditional methods}
Traditional detection methods are built on handcrafted features and shallow trainable architectures\cite{Zhao2019}. They are based on the truth that a license plate is in a rectangular shape with special colors and relatively length-fixed characters. Therefore, edge-based, color-based, texture-based, and character-based methods have been adopted. In \cite{Hsu2013}, edge clustering is formulated for solving plate detection. It is an edge-based method using Expectation Maximization (EM) to extract the regions of license plates. In \cite{Yuan2017}, a line density filter approach is proposed to extract candidate regions. Then, linear Support Vector Machine (SVM) is used to further classify the candidate regions based on color salience features. In \cite{Ashtari2014}, color pixels are used detect the location of a license plate based on a modified template-matching technique. The edge-based and color-based methods are usually fast enough for real-time applications. However, the edge and color features are sensitive to illumination conditions in unconstrained scenes. For example, edge features may be lost and colors may change under strong light. 

To overcome the problems coursed by illumination changes, \cite{YU2015114} proposes a robust texture-based method using wavelet transform and empirical mode decomposition (EMD) analysis to search for the location of a license plate. In \cite{Shemarry2020}, multi-level extended local binary pattern is used as an efficient detector. Compared with edge-based and color-based methods, this type of methods can alleviate the problems coursed by illumination changes and complex background, which is achieved by using more time-consuming processing modules. \cite{Setumin2010} presents a method that directly locates the alphanumeric characters of the car plate by using character features. In \cite{Li2013}, license plate is decomposed into several constituent characters. Then, conditional random field (CRF) and maximally stable extremal region (MSER) are used to extract and construct characters respectively. Character-based methods can adapt to various environmental factors to some extent. The disadvantage is that they can be easily affected by character-like background.

Summarily, traditional methods mainly extract low-level handcrafted features which have limited representation ability. Extra processing steps are always needed. Although complex pipelines are designed to handle unconstrained circumstances, performance can hardly be as good as that of deep learning-based methods.

\subsubsection{Deep learning-based methods}
Deep learning-based methods usually extract features and regress location parameters with deep neural networks. Most of them are built with supervised training. For feature extraction, most of the previous works adopt the commonly used backbones, such as VGG \cite{vgg2014}, DarkNet \cite{Redmon2016}, ResNet (50, 101, 152) \cite{He_2016_CVPR}, DenseNet \cite{Huang_2017_CVPR}, which are time-consuming. A few of works design lightweight networks for efficiency. For example, \cite{Silva_2018_ECCV} introduces a novel network with less than 20 convolutional layers. It is fast enough for real-time applications. 

For location, the generic object detection methods such as YOLO \cite{Redmon2016}, SSD \cite{Liu_2016_ECCV}, Faster-rcnn \cite{Ren2017}, can be used to get the bounding box of a license plate. The object detection methods can be roughly divided into three categories: anchor-based one-stage, two-stage and anchor-free methods. Anchor-based one-stage methods, such as YOLO and SSD, are frequently used for license plate detection due to their lower computation cost. For example, \cite{Laroca2018} presents a robust and efficient ALPR system using YOLO detector. In \cite{Xie2018}, a multi-directional detector based on modified YOLO framework is proposed for license plate images. Two-stage methods such as Faster-rcnn, have high detection accuracy but with low computational efficiency. \cite{Li2019} presents a network based on two-stage detection.  Candidate LP regions are detected by a Region Proposal Network (RPN). Then, a detection head is used for bounding box regression. Compared with anchor-based methods, anchor-free methods \cite{Law_2018_ECCV,centernet} avoid hyper-parameters related to the anchors and calculation of IoU (between ground truth boxes and anchor boxes), which make the training process more efficient and easier to train. In addition, some tricks can improve the performance. In \cite{Liu2019}, a hybrid cascade structure is designed to detect small and blurred license plates in complex scenes. In \cite{Zhang2019}, RNN is designed to improve the locating rate of license plates in complex scenes. In \cite{BJORKLUND2019134}, 4 corners of a license plate are predicted to enhance accuracy of detection. Summarily, deep learning-based methods are often more robust by training with numerous samples.

\subsection{License plate recognition}
Before deep learning technology was widely used, traditional methods for license plate recognition are typically based on character segmentation. Then, optical character recognition (OCR) methods are adopted to classify each character one by one. In [6], traditional methods were reviewed and compared from different perspectives. Benefiting from the development of deep learning technology and computing power in recent years, segmentation-free methods has gradually become the prevailing methods.

\subsubsection{Segmentation-based Methods}
Character segmentation is based on the truth that characters and background have obviously different colors in a license plate. Usually, the binary image of a license plate is needed to obtain the boundary of characters by horizontal pixel projection \cite{Wu2006, Goel2013, Gou2016, HENDRY201947}. Then, the problem is transformed into character recognition, which is a kind of image classification task. It should be noted that the accuracy of license plate recognition depends heavily on the quality of character segmentation.

Template matching and learning-based methods are two types of commonly used methods for plate recognition. In \cite{Goel2013}, segmented images are compared with predefined standard template and characters are recognized based on best match. In \cite{Gou2016}, hybrid discriminative restricted Boltzmann machine is trained for character classification. Traditional machine learning methods, such as hidden Markov Models (HMM) and Support Vector Machine (SVM), can also be used for classification \cite{Khan2018, Bulan2017}. In recent years, deep learning methods are frequently adopted for this problem. In \cite{HENDRY201947}, YOLO framework is used for both character segmentation and recognition. Similarly, an improved version of
YOLOv3 \cite{multinational} is employed to detect and classify each character in a license plate. In \cite{SELMI2020213}, Mask-rcnn \cite{He_2017_ICCV} method are adopted to segment each character from license plate area. Generally, template matching methods are simple to set up with fewer samples. On the contrary, learning based methods are often more robust by training with numerous samples. Generally, the learning based recognition methods, combined with character segmentation, are still widely used for LPR in some constrained scenarios.

\subsubsection{Segmentation-free Methods}
In order to avoid the uncertainty of character segmentation, segmentation-free methods have become a hot research topic in recent years. Segmentation-free methods usually take the license plate characters as a sequence directly. The recognition task can be transformed into sequence labeling problem. To solve this type of problem, both CNN-based and RNN-based methods have been adopted in different situations.

In \cite{lprnet}, CNN classifier is employed for license plate recognition directly. In \cite{Yang2018} CNN method is adopted to learn deep features. Then, a kernel-based ELM (Extreme learning machine) classifier is utilized for classification. In \cite{Xu_2018_ECCV}, multi-layer feature maps are extracted from different layers, resized to a fixed size by RoI pooling, and then concatenated to one feature map that used for character classification directly. Similarly, a multi-task deep convolutional network is proposed in \cite{8614318} to directly classify each character without segmentation. The reason why these methods are feasible is that the length of characters in a license plate is relatively fixed. Generally, the adaptability of CNN-based methods are relatively weak, and it is difficult to flexibly adapt to license plates of different lengths using a single model. Other methods such as Fully Convolutional Network (FCN) based methods \cite{Zhuang_2018_ECCV, Wu2017}, finding each character area by semantic segmentation, can also achieve relatively good results.

RNN-based methods are widely used in speech recognition, language modeling, machine translation, etc. RNN has natural advantages for sequence labeling due to their powerful connectionist mechanism. In \cite{WANG2018149}, bi-directional recurrent neural network (BRNN) is adopted for solving sequence labeling of license plate characters. Then, the recognition is accomplished by the BRNN and CTC \cite{Alex2006}. In \cite{LI201814} and \cite{Li2019}, Long Short-Term Memory (LSTM) net is trained to recognize the sequential features extracted from the whole license plate via CNN. In \cite{Liu2018}, RNN is designed to automatically recognize the characters of Chinese license plates. To improve performance in complex and unconstrained scenarios, a quality-aware plate stream recognition algorithm is proposed in \cite{EQ-LPR}. It is a video-based method that integrates efficient stream quality estimation and plate frame recommendation. However, RNN-based methods depend on the computations of the previous time step and therefore do not allow parallelization over every element in a sequence.

\subsection{End-to-end methods}
Recently, some end-to-end license plate detection and recognition methods have been proposed. Two sub-tasks are solved by single end-to-end trainable networks. Compared with two-step methods, end-to-end ones can reduce the propagation of error between detection and recognition models. In \cite{Xu_2018_ECCV}, multi-layer feature maps within a network are used for bounding boxes regression. Then, the recognition module extracts RoIs from shared feature maps and predicts the characters. This network is based on SSD. In \cite{Li2019}, RPN and RoI pooling are used for extracting region features. Then, detection branch is designed for bounding boxes regression, and recognition branch based on RNN is adopted for sequence labeling. This method is based on Faster-rcnn. The common characteristics of the two methods are the use of shared features, RoI pooling and multi-task learning. 

\begin{figure*}[!t]
	\centering
	\includegraphics[width=1.0\textwidth]{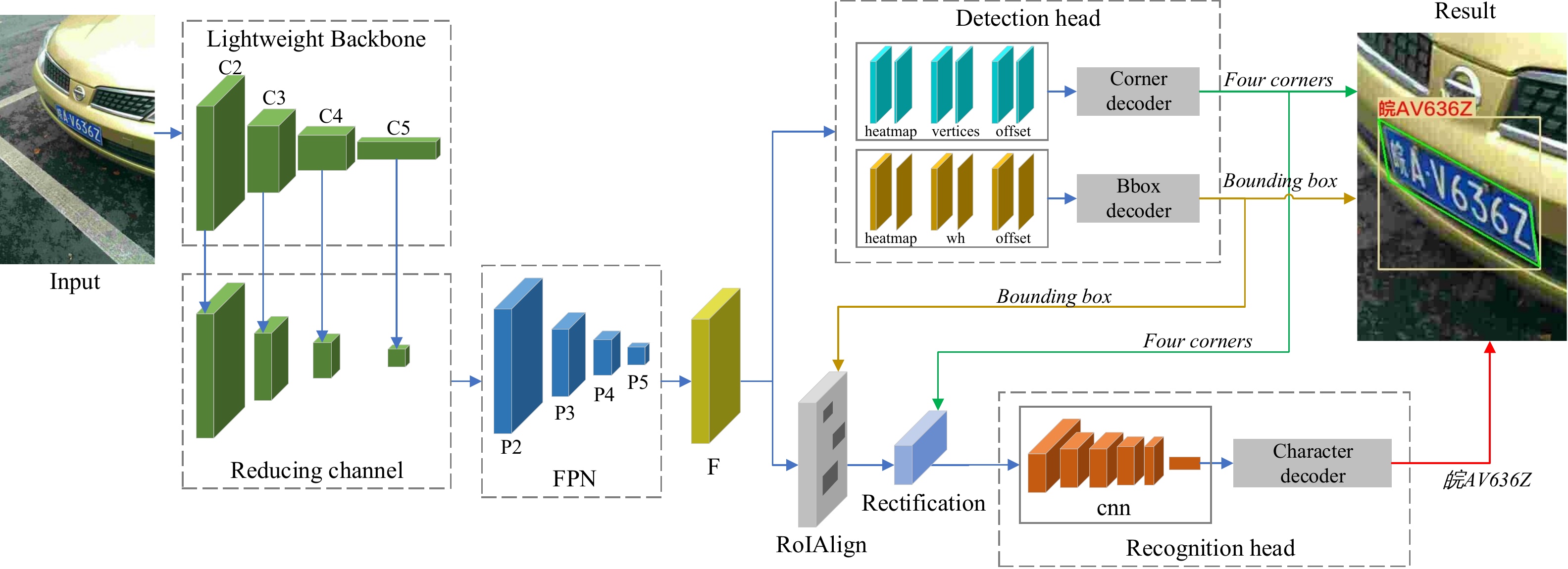}
	\caption{Proposed network for license plate location and recognition. It consists of feature extraction, location and recognition. FPN is employed to extract shared features for both classification and recognition. Then detection branch generates bounding box and corners that are used for RoIAlign and rectification respectively. Recognition branch is a lightweight CNN network designed for further feature extraction. Finally, recognition problem is solved sequence labeling. CTC method is adopted for character level classification directly. It is an end-to-end network that can work in real-time. 
	}
	\label{fig2}
\end{figure*}

Scene text detection and recognition \cite{Shi2017,Li_2017_ICCV,He2018} is similar to ALPR. In \cite{Li_2017_ICCV}, a unified network is proposed to localize and recognize text with a single forward pass. It is also based on Faster-rcnn. \cite{He2018} introduces a similar overall architecture that consists of detection and recognition branches. Some tricks are designed for better performance, such as text-alignment layer for arbitrary orientation detection and attention mechanism for recognition.

In summary, end-to-end methods are typically based on detection frameworks and with additional recognition branches. Through shared features extraction and multi-task training, the learned features become more informative, which can improve the overall performance to some extent.

\section{Proposed method}
\label{method}

\subsection{Overview}
The architecture of proposed network is depicted in Fig. \ref{fig2}. Different from the existing methods, our approach is a novel anchor-free network to solve license plate location and recognition tasks simultaneously. It consists of feature extraction, location branch and recognition branch. For shared feature extraction, we choose a lightweight backbone for efficiency. Then channel reducing and FPN are used. Inspired by the anchor-free method in \cite{centernet}, we design a detection head to regress bounding box and corners of a plate without NMS. Compared with single bounding box methods, the corners can mark the license plate more accurately. For recognition, RoIAlign\cite{He_2017_ICCV} is firstly used to cut and resize the shared feature maps by bounding box. Then, rectification is adopted with the help of corners.
We design the CNN-based recognition head to label and decode the character sequences without character segmentation. It is a real-time and end-to-end trainable framework that can work in unconstrained scenarios. 

\subsection{Network architecture}

\subsubsection{Feature extraction}
As shown in Fig. \ref{fig2}, We employ a lightweight model (ResNet-18 \cite{He_2016_CVPR}) as the backbone network for efficiency. Then, channel reducing strategy borrowed from \cite{Wang_2019_ICCV} is adopted to further reduce the computation cost. Specifically, 4 feature maps (denoted as $C_{2}, C_{3}, C_{4}, C_{5}$) generated by conv2, conv3, conv4, and conv5 stages of the backbone are used for channels reducing. It should be noted that 4 feature maps have strides of 4, 8, 16, 32 pixels with respect to the input image. Finally, the channel number of each feature map is reduced to 128 by using a $1\times1$ convolution layer (BN and ReLU are included). The reduced feature maps are denoted as $C_{2}^{'},C_{3}^{'},C_{4}^{'},C_{5}^{'}$.

Feature Pyramid Networks (FPN) \cite{Lin_2017_CVPR} is adopted to fuse different levels of feature maps from top to bottom. We firstly compute 4 new feature maps (named as $P_{2},P_{3},P_{4},P_{5}$), each with 128 channels. They are calculated by: 
\begin{equation}
\begin{aligned}
&P_{5}=C_{5}^{'}\\
&P_{4}=0.5\times C_{4}^{'} + 0.5\mathrm{Up}_{\times 2}(P_{5})\\
&P_{3}=0.5\times C_{3}^{'} + 0.5\mathrm{Up}_{\times 2}(P_{4})\\
&P_{2}=0.5\times C_{2}^{'} + 0.5\mathrm{Up}_{\times 2}(P_{3})
\end{aligned}
\end{equation} 
where $\mathrm{Up}_{\times 2}(\cdot )$ refers to 2 times up-sampling. The nearest neighbor interpolation algorithm is used for up-sampling. To further fuse the semantic features between low and high levels, we use the similar method in \cite{Wang_2019_CVPR} to calculate the feature map $F$. It is described as follows:
\begin{equation}
F=P_{2}\cup \mathrm{Up}_{\times 2}P_{3}\cup \mathrm{Up}_{\times 4}P_{4}\cup \mathrm{Up}_{\times 8}P_{5}
\end{equation}   
where $\cup$ refers to the channel concatenation, $\mathrm{Up}_{\times 4}(\cdot )$ and $\mathrm{Up}_{\times 8}(\cdot )$ refer to 4 and 8 times up-sampling respectively. Then, a $3\times3$ convolution layer (with BN and ReLU) is used to reduce the channels number of $F$ to 128. 

In summary, we provide a lightweight feature extraction network based on ResNet-18 and FPN. For up-sampling in the network, we use a simple nearest neighbor interpolation algorithm for efficiency. The size of output shared feature maps is $B\times\frac{1}{4}H\times\frac{1}{4}W\times128$, where $B$ is the batch size, and $H, W$ refer to the input image size. 

\subsubsection{Location network}

\begin{figure}[!t]
	\centering
	\includegraphics[width=0.45\textwidth]{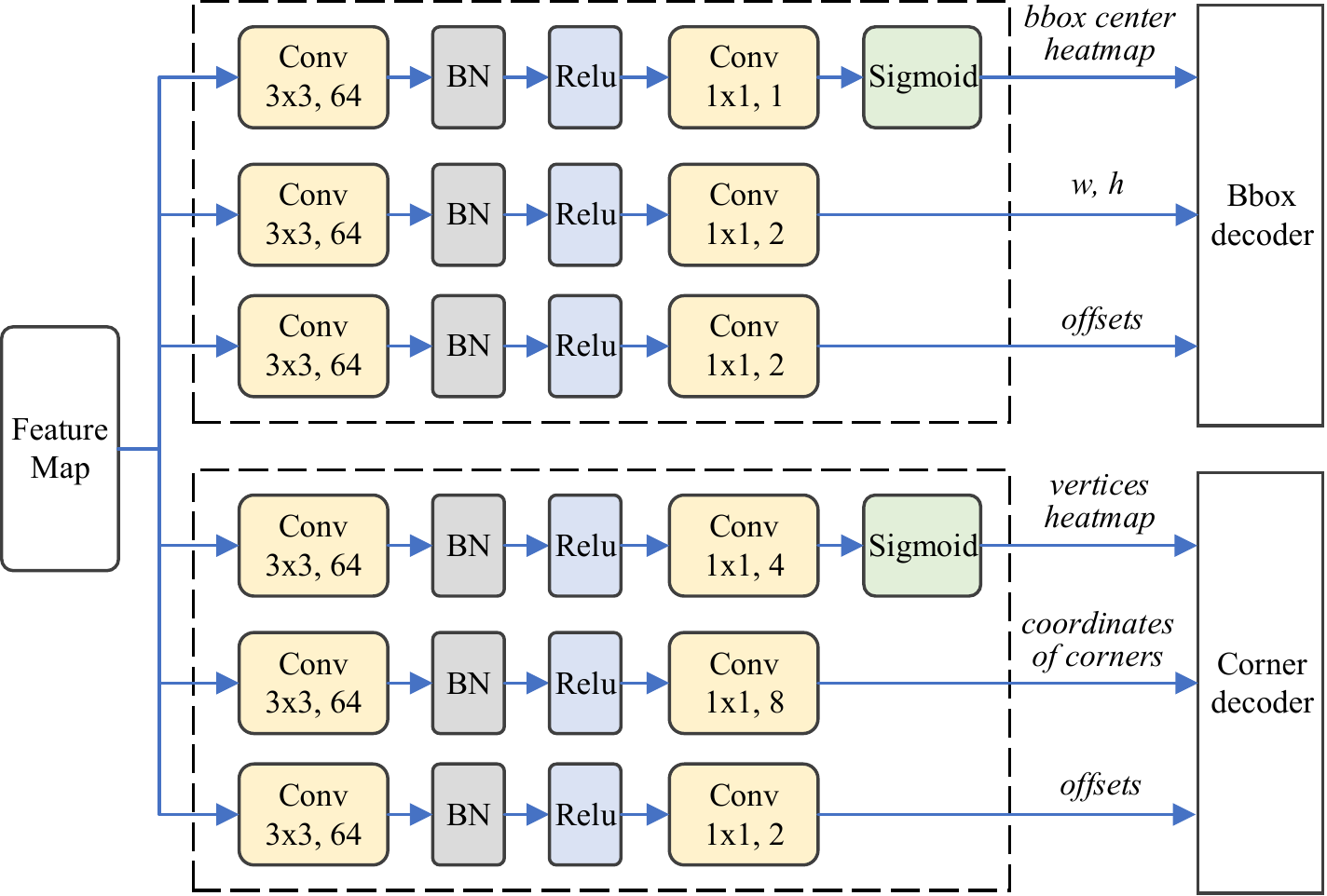}
	\caption{Proposed detection head. It consists of 6 sub-branches, 3 for bounding box regression and 3 for corner regression. We use a center point, $w, h$ and offsets to represent a bounding box, use the relative positions to mark 4 corners of a license plate. Bounding box decoder and corner decoder are designed to decode the box and corners respectively.} 
	\label{fig3}
\end{figure}

In order to adapt unconstrained scenarios, we designed a detection head to accurately locate a license plate by bounding box combined with four corners. Inspired by \cite{centernet}, the bounding box of a license plate is denoted by a center point and $w, h$ parameters. As illustrated in Fig. \ref{fig3}, there are 6 sub-branches, 3 for bounding box regression and the other 3 for corner regression. For each sub-branch, a $3\times3$ convolutional layer and a $1\times1$ convolutional layer are used. The channel number of the $3\times3$ convolutional layer is set to 64 for all sub-branches. The $1\times1$ convolutional layers are used to format the output feature maps. Different types of sub-branches have different channel numbers. The sub-branches with Sigmoid activation are heatmaps used to predict the centers of bounding boxes and corners. For bounding boxes, $w, h$ are regressed by a sub-branch with 2 output channels. For corners, relative positions are predicted by the sub-branch with 8 output channels. The left sub-branches are used to calculate the offsets caused by feature maps down-sampling (from $H\times W$ to $\frac{1}{4}H\times\frac{1}{4}W$). The prediction of corners is based on the truth that there are obvious texture and shape features on the four corners. We treat the corners as key points of a license plate, and regress the positions relative to the center point. The proposed detection head is depicted in Fig. \ref{fig3}. Although we predict four corners which can accurately represent a license plate, the bounding box is also useful for plate location and rectification. Firstly, it is used to help us determine which plate a corner belongs to. When there is more than one plate in an image, we calculate the distances between corners and bounding boxes. The closest corners are attached to the bounding box. Secondly, it is also employed at RoIAlign stage to crop features for rectification.

Since the proposed location branch is a dense prediction network, each point of the heatmap is a prediction result. For bounding boxes decoding, peaks in the heatmap are extracted as center points. Then, the $w, h$ and offsets for each center point can extracted from the other two sub-branches. For corner decoding, corners are strictly divided into four categories (left-top, right-top, left-down, right-down). Then, peaks in the heatmap are extracted for each category. After finding all bounding boxes and corners, we use a simple distance measure between bounding boxes centers and corners to determine which box the corners belongs to.

\subsubsection{RoIAlign and rectification}
RoIAlign \cite{He_2017_ICCV} is employed to crop and resize the shared feature maps with the bounding boxes decoded by location network. As we know, a license plate is rectangular. We set $h=32, w=96$ to resize the cropped feature. Then, the size of feature maps after RoIAlign is changed to $B_{r}\times 32 \times 96 \times 128$, where $B_{r}=B\times M$ is the new batch size for recognition. $M$ is the maximum number of license plate instances per image. Compared with RoI pooling, RoIAlign can generate feature maps with more accurate boundaries.  

Rectification is designed to unwrap the feature maps with perspective distortion. In unconstrained scenes, the location of a license plate denoted by a single bounding box is inaccurate, because a lot of invalid pixels outside the license plate area are included. In this situation, the features after RoIAlign are distorted and noisy, which will affect the performance of recognition. To overcome this problem, we use four corners to accurately location a license plate. Specifically, projective transformation is used to map one vector space formed by corners into another one defined by the corresponding bounding box. The transformation can be done by the simple multiplication:
\begin{equation}
\begin{bmatrix} x^{'}\\ y^{'}\\ 1 \end{bmatrix}=\begin{bmatrix}a1 & a2 & b1\\ a3 & a4 & b2\\ c1 & c2 & 1\end{bmatrix}\times \begin{bmatrix}x\\ y\\ 1\end{bmatrix}
\end{equation}   
where $x,y$ are the coordinates of a point within the quadrangle connected by the 4 corners, $x^{'},y^{'}$ are the coordinates of the transformed point, $\begin{bmatrix}a1 & a2\\ a3 & a4\end{bmatrix}$ is a rotation matrix, $\begin{bmatrix}b1\\ b2\end{bmatrix}$ is the translation vector, $\begin{bmatrix}c1& c2\end{bmatrix}$ is the projection vector. Firstly, we calculate the transformation matrix using the 4 corners and the feature size after RoIAlign. Then, we transform the points to new coordinates. After rectification, the size of feature maps is not changed.

\subsubsection{Recognition network}

\begin{figure}[!t]
	\centering
	\includegraphics[width=0.4\textwidth]{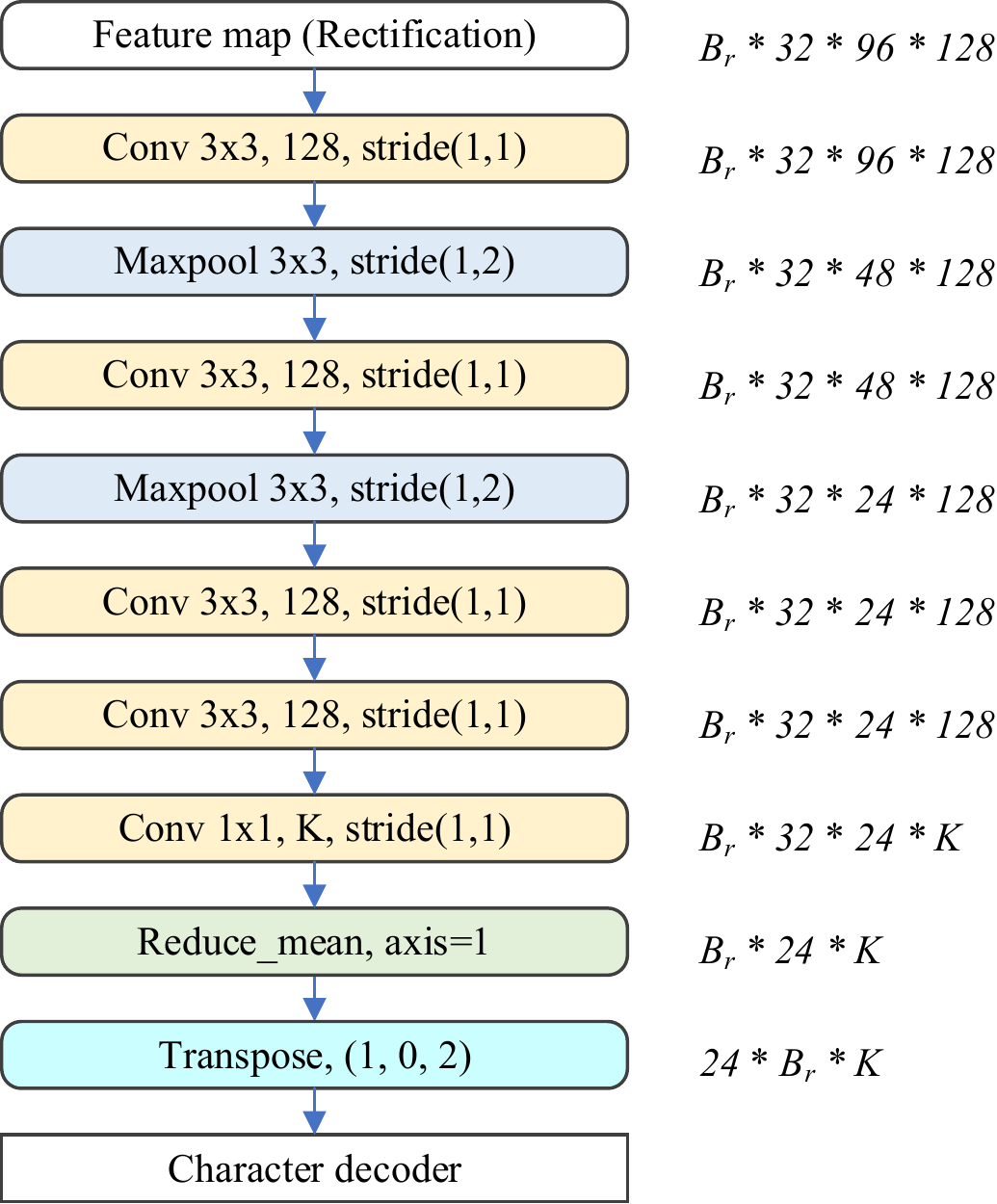}
	\caption{Proposed recognition head. It consists of 5 convolutional layers, 2 Max pooling layers and a character decoder module. The size of feature maps after processing at each layer is denoted on the right side.}
	\label{fig4}
\end{figure}

We treat license plate recognition as sequence labeling without character segmentation. The goal of sequence labeling is to assign sequences of labels. Since there is no correlation between the characters, we use a simple CNN network to simulate temporal classification for efficiency. Compared with segmentation-based methods, temporal classification data can be weakly labeled with nothing but the target sequences. As shown in Fig. \ref{fig4}, we design a small network with only 5 convolutional layers and 2 Max pooling layers for further feature extraction. The last convolutional layer with $1\times1$ kernel and $K$ channels is used to extract the final features. $K=C+1$, where $C$ is the number of character categories. The size of feature maps after computing by each layer is denoted on the right side in Fig. \ref{fig4}. The size of the final features is $24\times B_{r} \times K$, where 24 is the number of time steps. The character decoder is designed to decode characters from the final features. Since we train the network using CTC, the decoding process is to retrieve the most probable labeling $z^{*}$.  
\begin{equation}
z^{*}=argmax P(z|x) \label{eq1}
\end{equation}
where $z$ is the labeling path, $x$ is the input sequence to decode. To solve the problem in (\ref{eq1}), we choose Beam Search method, since there is a trade-off between search time and accuracy.

\subsection{Optimization}
\subsubsection{Label assignment}

\begin{figure}[!t]
	\small
	\centering{\includegraphics[width=0.48\textwidth]{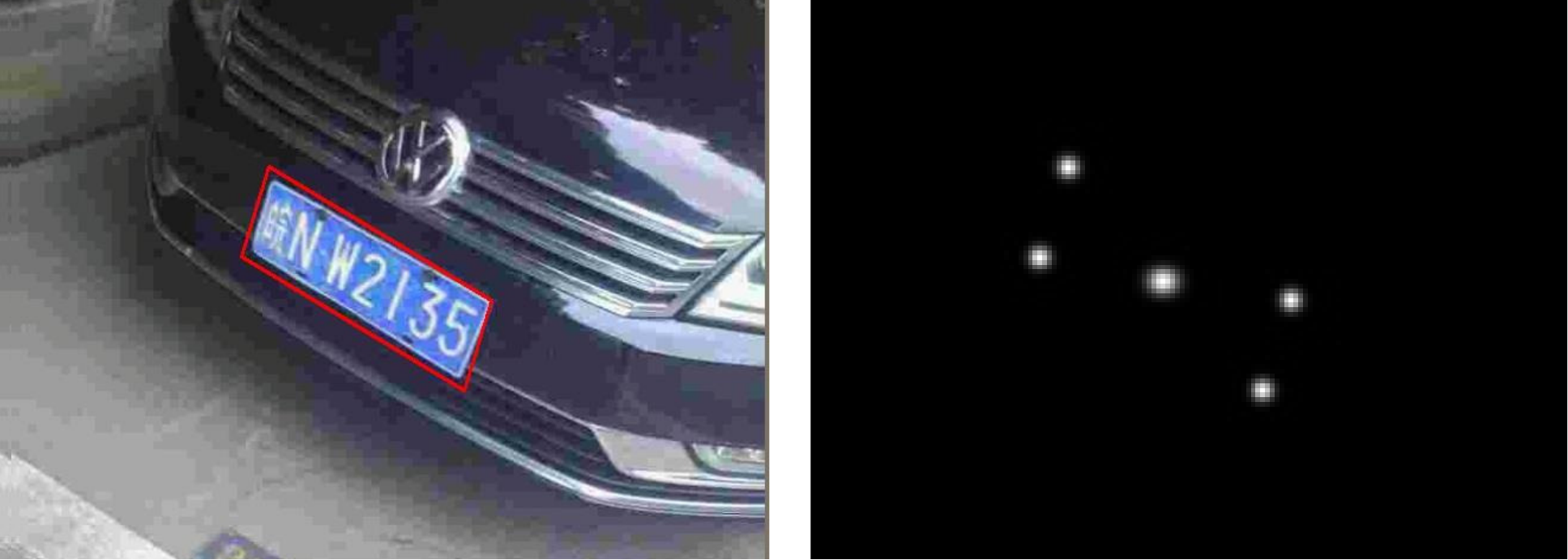}}
	\caption{Label assignment for a license plate. The center and corners are denoted by values ($[0,1]$) with Gaussian distributions. The center points of these distributions are set to 1. The right image is a fusion of five labeled maps. \label{fig5}}
\end{figure}

For location, the network need to be given the positions of corners and center of license plates. For each center, $w, h$ and offsets are attached. And for each corner, relative coordinates $x, y$ and offsets are appended. As shown in Fig. \ref{fig5}, the center and corners are denoted by values ($[0,1]$) with Gaussian distributions. The center points of these distributions is set to 1. The heatmap $Y\in [0,1]^{\frac{1}{4}H\times \frac{1}{4}W}$ is generated with a Gaussian kernel \cite{centernet}:
\begin{equation}
Y = \mathrm{exp}(-\frac{(x-x_{c})^{2}+(y-y_{c})^{2}}{2\sigma _{c}^{2}}) \label{eq5}
\end{equation}
where $(x_{c}, y_{c})$ is coordinates of a center or a corner, $\sigma _{c}$ is a standard deviation. We set $\sigma _{c}=\frac{1}{3}r$ according to \cite{Law_2018_ECCV}. $r$ is a predefined radius. We set $r=0.4\times \mathrm{min}(w, h)$ for center and $r=0.2\times \mathrm{min}(w, h)$ for corners. We think that the radius $r$ should be proportional to the size of plate, because small object should be represented by small heatmap. So, we choose $\mathrm{min}(w, h)$ as a basis. Then, we set the factor to $0.4$ and $0.2$ for center and corners respectively in consideration of the size of heatmap. We use a smaller factor for corners due to the fact that corners should be located more precisely. For recognition, the character sequences are encoded as sequences of class ids.

\subsubsection{Loss functions}
The training task includes two parts: location and recognition. So the loss function can be formulated as: 
\begin{equation}
L=L_{d}+\lambda L_{r} \label{eq6}
\end{equation}
Here $L_{d}$ is the location loss, $L_{r}$ is the recognition loss. We set the weight factor $\lambda=10$ in our experiments. The location loss $L_{d}$ is described in:
\begin{equation}
L_{d}= L_{c}^{b}+L_{wh}^{b}+\beta L_{off}^{b} + L_{c}^{c}+ L_{wh}^{c}+\beta L_{off}^{c}\label{eq7}
\end{equation}
where $L_{c}^{b}$ and $L_{c}^{c}$ refer to the position losses for centers and corners respectively, $L_{wh}^{b}$ is the $w, h$ loss for bounding boxes, $L_{wh}^{c}$ is the relative coordinates loss for corners, $L_{off}^{b}$ and $L_{off}^{c}$ are offsets losses for bounding boxes and corners respectively. The weight factor $\beta$ is set to $0.05$. Specifically, $L_{c}^{b}$ and $L_{c}^{c}$ are based on the focal loss\cite{Lin_2017_ICCV}. They can be formulated as $L_{c}$:
\begin{equation}
\begin{aligned}
&L_{c}^{pos}=-\sum \mathrm{log(\hat{Y})\times (1-\hat{Y})^{2}}\\
&L_{c}^{neg}=-\sum \mathrm{log(1-\hat{Y})\times \hat{Y}^{2}\times(1-Y)^{4}}\\
&L_{c} = \left\{\begin{matrix}
\frac{1}{N^{pos}}(L_{c}^{pos}+L_{c}^{neg}) & \mathrm{if}N^{pos}>0\\ 
L_{c}^{neg}& \mathrm{else} 
\end{matrix}\right. \label{eq8}
\end{aligned}
\end{equation}
where $Y$ is the ground truth value, $\hat{Y}$ is the predicted value, $N^{pos}$ is the number of positive samples in an image. $L_{c}^{pos}$ and $L_{c}^{neg}$ refer to positive and negative losses respectively. $L_{wh}^{b}$, $L_{wh}^{c}$, $L_{off}^{b}$, $L_{off}^{c}$ are based on $L1$ loss. They can be formulated as $L_{l1}$:
\begin{equation}
L_{l1}=\frac{1}{N}\sum \mathrm{abs}(Y-\hat{Y})
\label{eq9}
\end{equation}
The standard CTC loss is used as the recognition loss $L_{r}$. It is given by:
\begin{equation}
L_{r} =-\sum_{(x,z)\in S}lnP(z|x) \label{eq10}
\end{equation}
where $P(z|x)$ represents the conditional probability of the output target sequence $z$ through the input $x$. $S$ is defined as the training dataset. In this paper, the length of the final feature for decoding is set to $L=24$. Because we use the CNN output to decode directly, the length $L$ is used to indicate the length of time. Defining $p_{\pi_{t}}^{t}$ as the predicted probability of sequence $\pi_{t}$ at time $t$, $P(z|x)$ can be formulated as: 
\begin{equation}
P(z|x)=\prod_{t=1}^{L}p_{\pi_{t}}^{t} \label{eq11}
\end{equation}

\subsection{Post-processing}

For location, the results from bounding box decoder and corner decoder should be restored to original image size by $4$ times multiplication. Then, invalid values are clipped. For recognition, since the license plate is a special character sequence with certain specifications, we filter the decoding sequences by setting specific rules. That is to say, after character decoding, the results that do not meet to the rules are discarded firstly. We formulate license plate recognition rules consisting of two parts: number of characters rule and character subset rule. The first rule is adopted to filter out the result sequences whose length do not meet the requirements. The second rule defined a valid subset for each character of the license plate. The result sequences will be filtered out if any character is not in its subset. We do post-processing after the Beam Search. The first sequence that meets the predefined rules is the final result. It should be noted that post-processing is optional.

\section{Experiments}
\label{experiment}
In this section, we experimentally evaluate the proposed approach on several public datasets, including CCPD dataset\cite{Xu_2018_ECCV}, AOLP dataset\cite{Hsu2013} and PKU vehicle dataset\cite{Yuan2017}. For detection performance metrics, the Intersection over Union (IoU) between predicted box and ground truth box is adopted to judge if the detection result is matched to the ground truth. The threshold $\lambda$ of IoU is generally set to $\lambda=0.5$. When IoU $>\lambda$, the predicted box is correct. For end-to-end performance metrics, 
detection and recognition should be considered simultaneously. Following the general end to-end evaluation protocol used in \cite{Li2019}, we calculate the IoU and recognition results. When IoU $>\lambda$ and recognition result is matched for a license plate, the prediction is correct. 

\subsection{Datasets}
\subsubsection{CCPD dataset}
CCPD contains large-scale images with detailed annotations including LP number, LP bounding box, four vertex locations and other information. The resolution ($w\times h$) of each image is $720\times 1160$. Specifically, it is divided into 9 sub-datasets including Base (Common photos with license plates), DB (Illuminations are dark, uneven or extremely bright), FN (The distance from LP to shooting location is relatively far or near), Rotate (Images taken by rotated cameras), Tilt (Images taken by rotated cameras), Blur (Blurry due to hand jitters), Weather (Images with rain, snow or fog), Challenge (The most challenging images) and NP (Images without a LP) \cite{Xu_2018_ECCV}. There are two versions of the dataset. The old version was released in $2018$. Baseline performances in \cite{Xu_2018_ECCV} are based on this version. In $2020$, the authors released a new version which is more challenging than before with over $300k$ images and refined annotations. In addition, standardized split file is available in the new version.

\subsubsection{AOLP dataset}
AOLP (application-oriented license plate dataset) consists of 2049 images of Taiwan license plates. It is split into three categories: AC (access control, 681 images), LE (law enforcement, 757 images), and RP (road patrol, 611 images). Specifically, AC refers to the cases that a vehicle passes a fixed passage at a reduced speed or with a full stop, LE refers to the cases that a vehicle is captured by a roadside camera, RP refers to the cases that a vehicle are captured by another moving vehicle \cite{Hsu2013}. The bounding box of each license plate is given by the annotation.

\subsubsection{PKU vehicle dataset}
The PKU vehicle dataset contains 3977 images captured under diverse conditions and in diverse environments. It is divided into 5 groups ( G1, G2, G3, G4, G5) corresponding to different conditions (such as scene, weather, time, distance) \cite{Yuan2017}. The annotation of each image is only a mask denoting each license plate by a rectangle, which is equivalent to a bounding box. Accordingly, it can be only used for license plate detection. 

\subsection{Implementation details}
Our network is trained with Adam \cite{Adam} optimizer on a GeForce RTX 2080Ti GPU with 11GB memory. Learning rate is set to $1\times 10^{-3}$ initially, and decayed using exponential strategy. The training batch size is set to $16$. To prevent over fitting, we use the data augmentation strategies including color jitterbugging, random cropping, random scaling, random translating and random rotation. For CCPD and AOLP datasets, we train the model end-to-end. For PKU vehicle dataset, the recognition branch is not trained, since the dataset is only designed for detection task. The backbones (ResNet-18, ResNet-50) pre-trained on ImageNet \cite{ImageNet} are used for experiments.

\subsection{Performances on CCPD dataset}

For old version of dataset, we divide Base sub-dataset into two equal parts, just as the baseline method does \cite{Xu_2018_ECCV}. For new version, we use the default split file provided by \cite{Xu_2018_ECCV} to extract training and testing subsets for fair comparison. Specifically, images in Base sub-dataset is split to training and validation sets. We train the model for 50 epochs on training set which has $10k$ images. The initial learning rate is set to $1\times 10^{-3}$, and is divided by 10 at 20 and 35 epochs. Sub-datasets (DB, Blur, FN, Rotate, Tilt, Challenge) are exploited for testing. Validation set is also used for testing as Base sub-dataset. For fair comparison, we follow the metric provided by \cite{Xu_2018_ECCV}. Specifically: (1) As each image in CCPD contains only a single license plate, recall is not considered (only concentrate on precision). Detectors are allowed to predict only one result for each image. (2) For this dataset, IoU is set to $\lambda=0.7$. Detection and Recognition performance are compared with state-of-art methods respectively using the same models. The resolution of images used for the following experiments is always set to $640\times 1024$ after letterbox resizing (Keep the original aspect ratio in the resized images).

\subsubsection{Detection}
\begin{table*}[!ht]
	\renewcommand{\arraystretch}{}
	\caption{Detection performances (as percentages) on different sub-datasets. We compare our methods with six state-of-art methods including SSD, Faster-rcnn, YOLOv3, TE2E, WPOD and RPnet. Results demonstrate that our methods are more accurate on most of the sub-datasets. In addition, our methods can run in real-time with 36 fps for Resnet-18 and 26 fps for Resnet-50 respectively. Results marked with * are based on the old version of CCPD dataset.}
	\label{tab1}
	\centering
	\begin{tabular}{lcccccccccc}
		\toprule	
		Method &Base &Rotate &Tilt &FN &DB &Weather &Challenge &Avg &Speed (FPS)\\	
		\midrule
		SSD\cite{Liu_2016_ECCV}*   &99.1 &95.6 &94.9 &84.7 &89.2 &83.4 &93.1 &94.4 &40\\
		Faster-rcnn\cite{Ren2017}* &98.1 &91.8 &89.4 &83.7 &92.1 &81.8 &83.9 &92.9 &15\\
		TE2E\cite{Li2019}*         &98.5 &95.1 &94.5 &83.8 &91.7 &83.6 &93.1 &94.2 &3\\
		RPnet\cite{Xu_2018_ECCV}*  &99.3 &94.7 &93.2 &85.3 &89.5 &84.1 &92.8 &94.5 &\textbf{61}\\
		YOLOv3\cite{yolov3}       &98.9 &95.7 &94.2 &86.3 &93.2 &85.2 &92.7 &94.8 &52\\
		WPOD\cite{Silva_2018_ECCV}&99.1 &97.9 &96.0 &86.1 &87.7 &95.4 &92.9 &95.1 &44\\
		Ours (ResNet-18)*           &99.5 &98.4 &95.9 &91.5 &90.6 &\textbf{99.1} &93.3 &97.5 &36\\
		Ours (ResNet-50)* &\textbf{99.7} &\textbf{98.5} &\textbf{96.2} &\textbf{94.5} &\textbf{92.7} &99.1 &93.5 &\textbf{97.8} &26\\
		Ours (ResNet-18)           &99.5 &98.3 &96.1 &92.6 &91.4 &98.6 &93.9 &97.3 &36\\
		Ours (ResNet-50) &99.6 &98.5 &95.9 &94.5 &92.4 &98.7 &\textbf{94.1} &97.6 &26\\
		\bottomrule
	\end{tabular}
\end{table*}

We compare the proposed method with other state-of-art methods including SSD \cite{Liu_2016_ECCV}, Faster-rcnn \cite{Ren2017}, YOLOv3 \cite{yolov3}, TE2E \cite{Li2019}, WPOD \cite{Silva_2018_ECCV} and RPnet \cite{Xu_2018_ECCV}. Results marked with * are based on the old version of CCPD dataset, while others are based on the new version. The detection performances are shown in Table~\ref{tab1}. It should be noted that the experimental results of the three methods (SSD, Faster-rcnn, TE2E) are provided by \cite{Xu_2018_ECCV} as baseline performances. YOLOv3 model is trained and tested with image size setting to $608\times 608$. In addition, the results of WPOD method are from the re-implemented version of us, since there is no performance on CCPD dataset in the original paper. We use two backbones (ResNet-18, ResNet-50) in proposed method. The IoU calculation is only based on the bounding box results. The location results of corners are not used for detection. The methods for comparison are of different types including one-stage anchor-based method (SSD, YOLOv3), two-stage method (Faster-rcnn), end-to-end methods (TE2E, RPnet). The comparative results indicate that our approaches are more accurate on most of the sub-datasets, especially on Rotate, FN and Weather sub-datasets. 

Specifically, our methods (ResNet-18) achieve average performance of 97.5\% and 97.3\%, which is at least $2\%$ higher than other methods. In addition, our models can run in real-time at presentable speeds ($36fps$ for ResNet-18 and $26fps$ for ResNet-50). We find that WPOD method and our methods achieve better performances than the others. It is partially because more information (corners) are used for training. 

\subsubsection{Recognition}
We compare the proposed method with several complete solutions (detection and recognition) including SSD \cite{Liu_2016_ECCV} + HC \cite{Holistic}, Faster-rcnn \cite{Ren2017} + HC \cite{Holistic}, TE2E \cite{Li2019}, WPOD \cite{Silva_2018_ECCV} + OCR \cite{Silva_2018_ECCV} and RPnet \cite{Xu_2018_ECCV}. The recognition performances are shown in Table~\ref{tab2}. Results marked with * are based on the old version of CCPD dataset, while others are based on the new version. The experimental results of SSD + HC, Faster-rcnn + HC, TE2E and RPnet are borrowed from the baseline of dataset \cite{Xu_2018_ECCV}. The solution (WPOD + OCR) from \cite{Silva_2018_ECCV} is a re-implemented version for experiments.

\begin{table*}[!t]
	\renewcommand{\arraystretch}{}
	\caption{Recognition performances (as percentages) on different sub-datasets. We compare our methods with five solutions including SSD + HC, Faster-rcnn + HC, TE2E, WPOD + OCR and RPnet. The results demonstrate that our methods are more accurate on most of the sub-datasets. Results marked with * are based on the old version of CCPD dataset.}
	\label{tab2}
	\centering
	\begin{tabular}{lcccccccccc}
		\toprule	
		Method &Base &Rotate &Tilt &FN &DB &Weather &Challenge &Avg &Speed (FPS)\\	
		\midrule
		SSD\cite{Liu_2016_ECCV} + HC\cite{Holistic}* &98.3 &88.4 &91.5 &\textbf{95.9} &96.6 &87.3 &83.8 &95.2 &35\\
		Faster-rcnn\cite{Ren2017} + HC\cite{Holistic}* &97.2 &82.9 &87.3 &90.9 &92.1 &85.5 &76.3 &92.8 &13\\
		TE2E\cite{Li2019}*  &97.8 &87.9 &92.1 &94.5 &94.8 &86.8 &81.2 &94.4 &3\\
		RPnet\cite{Xu_2018_ECCV}*  &98.5 &90.8 &92.5 &94.3 &\textbf{96.9} &87.9 &85.1 &95.5 &\textbf{61}\\
		WPOD + OCR\cite{Silva_2018_ECCV}&98.7 &94.5 &95.4 &85.2 &86.5 &94.8 &91.2 &94.6 &31\\
		Ours (ResNet-18)*   &99.3 &97.9 &95.5 &93.2 &92.9 &98.8 &92.4 &97.2 &36\\
		Ours (ResNet-50)* &\textbf{99.5} &\textbf{98.2} &\textbf{95.9} &93.7 &93.3 &\textbf{98.9} &92.9 &\textbf{97.6} &26\\
		Ours (ResNet-18)   &99.2 &97.9 &95.5 &92.3 &91.1 &98.2 &92.9 &96.9 &36\\
		Ours (ResNet-50) &99.4 &98.2 &95.6 &93.9 &92.0 &98.4 &\textbf{93.4} &97.3 &26\\
		\bottomrule
	\end{tabular}
\end{table*}

The results indicate that our methods outperform these ALPR solutions on most of the sub-datasets. For FN and DB sub-datasets, we achieve lower accuracy. That is probably because we train the model on the mixed dataset and do not augment the training data by sampling several times to increase the sub-training sets, compared with the solutions achieving better performances. Additionally, DB sub-dataset has a different color distribution, which can also reduce the accuracy. Our methods (ResNet-18) achieve average performance of 96.9\% and 97.2\%, which is at least $1.4\%$ higher than other methods. The reason why the gap ($1.4\%$) looks small is that simple samples (Base sub-dataset) occupy a large proportion. Compared with TE2E, our methods (ResNet-18) can run 10 times faster. Although RPnet is faster, our method is more accurate especially for Rotate, Tilt and Challenge cases. That is to say, the proposed method is more capable for unconstrained scenarios. Example results are shown in Fig. \ref{fig6} (a).

\subsection{Performances on AOLP dataset}

\begin{table}[!t]
	\renewcommand{\arraystretch}{}
	\caption{Detection and recognition performances (as percentages) on RP sub-dataset. Speed is valued in FPS. Experimental results demonstrate the efficiency and accuracy of the proposed method.}
	\label{tab3}
	\centering
	\resizebox{0.48\textwidth}{!}{
	\begin{tabular}{lccc}
		\toprule	
		Method &Detection &Recognition &Speed\\
		\midrule
		Hsu\cite{Hsu2013} &94.00 &85.70 &3.1  \\
		LSTMs\cite{LI201814} &95.58 &83.38 &$<1$  \\
		TE2E\cite{Li2019} &98.85 &83.63 &2.5  \\
		Segmentation\cite{Zhuang_2018_ECCV}  &- &99.02 &-\\
		WPOD+OCR\cite{Silva_2018_ECCV} &98.54 &93.29 &38.5  \\
		WPOD+OCR\cite{Silva_2018_ECCV}* &98.54 &98.36 &38.5  \\
		Multinational\cite{multinational} &\textbf{100} &\textbf{99.51} &34.5  \\
		Ours (ResNet-18) &98.20 &93.78 &\textbf{47.6} \\
		Ours (ResNet-18)* &99.36 &98.82 &47.6 \\
		\bottomrule
	\end{tabular}}
\end{table}

Because AOLP dataset has no annotation for corners, we can't train a model directly without complete labels. By undertaking the analysis of the dataset, we find that almost all the images from AC sub-dataset are taken with no tilt angle, a few images from LE sub-dataset are taken by tilt cameras. On the contrary, most of the images from RP sub-dataset are taken by tilt cameras. In order to generate labels for corners, we choose AC and LE sub-datasets as the training dataset, from where the invalid images (taken by tilt cameras) are removed. Firstly, the bounding box of each license plate is used to generate the coordinates of four corners based on the truth that the bounding box can approximately represent the accurate boundary. Then, we randomly rotate the images and labels with degree from $-45^{\circ}\sim 45^{\circ}$. New corners are calculated by the rotation matrix and the new bounding boxes are generated using the new corners. It should be noted that the generated labels for corners are not always accurate due to the approximate representation. RP sub-dataset is used for test. The resolution of the images is set to $512\times 640$ after letterbox resizing.

We compare our method with state-of-the-art methods on this dataset, including Hsu \cite{Hsu2013}, LSTMs \cite{LI201814}, TE2E \cite{Li2019}, Segmentation \cite{Zhuang_2018_ECCV}, WPOD + OCR \cite{Silva_2018_ECCV} and Multinational\cite{multinational}. Results are shown in Table~\ref{tab3}. It should be noted that the recognition results are the overall results from input images to output license plate characters, and the speed is also for overall. Compared with TE2E, our method can increase the recognition performance by $10\%$. And the computational speed is also significantly improved. For WPOD + OCR, the detection performance and speed are based on the re-implemented version. WPOD + OCR* means synthetic data is added for training. The addition of synthetic data improves the accuracy to from $93.29\%$ to $98.36\%$. To achieve better performance, we also add synthetic data to train the proposed end-to-end model, denoted as Ours (Resnet-18)*. With more training data, recognition accuracy is increased from $93.78\%$ to $98.82\%$. In contrast, our method can handle the cases with rotated and tilted license plates. Although some methods \cite{Zhuang_2018_ECCV} and \cite{multinational} can achieve better performance without using additional dataset for training, character-level annotations are always needed. We argue that the proposed segmentation-free recognition method needs more data for training compared with the segmentation-based ones. Given enough data, the proposed method can achieve comparable performance. Example results are shown in Fig. \ref{fig6} (b).

\subsection{Performances on PKU vehicle dataset}
PKU vehicle dataset is only used for detection task. We can only get the bounding boxes from denotations. Similarly, we generate labels for corners to train our model. We just remove the recognition loss when training, due to the lack of ground truth for characters. We randomly choose $50\%$ images for training and the rest for testing. ResNet-18 is used as the backbone. We use the pre-trained model on CCPD dataset to fine tune on this training set. Similarly, the WPOD model is also pre-trained on CCPD dataset and fine tune on this training set. The resolution of images is set to $512\times 640$ after letterbox resizing.

\begin{table}[!t]
	\renewcommand{\arraystretch}{}
	\caption{Detection performances (as percentages) on PKU vehicle dataset. Speed is valued in FPS. We compare our approach with three different methods. Experimental results demonstrate the efficiency and accuracy of our method.}
	\label{tab4}
	\centering
	\resizebox{0.48\textwidth}{!}{
	\begin{tabular}{lccccccc}
		\toprule	
		Method &G1 &G2 &G3 &G4 &G5 &Avg &Speed\\
		\midrule
		Li\cite{Li2013} &98.89 &98.42 &95.83 &81.17 &83.31 &91.52 &1.5 \\
		Yuan\cite{Yuan2017} &98.76 &98.42 &97.72 &96.23 &97.32 &97.69 &23.8  \\
		TE2E\cite{Li2019} &\textbf{99.88} &99.86 &\textbf{99.87} &\textbf{100} &99.38 &\textbf{99.80} &3.3 \\
		WPOD\cite{Silva_2018_ECCV}&99.75 &99.86 &99.64 &96.65 &99.47 &99.67 &\textbf{55.6}  \\
		Ours &99.75 &\textbf{100} &99.73 &99.65 &\textbf{99.65} &99.80 &47.6 \\
		\bottomrule
	\end{tabular}}
\end{table}

\begin{figure*}[!t]
	\centering{\includegraphics[width=1.0\textwidth]{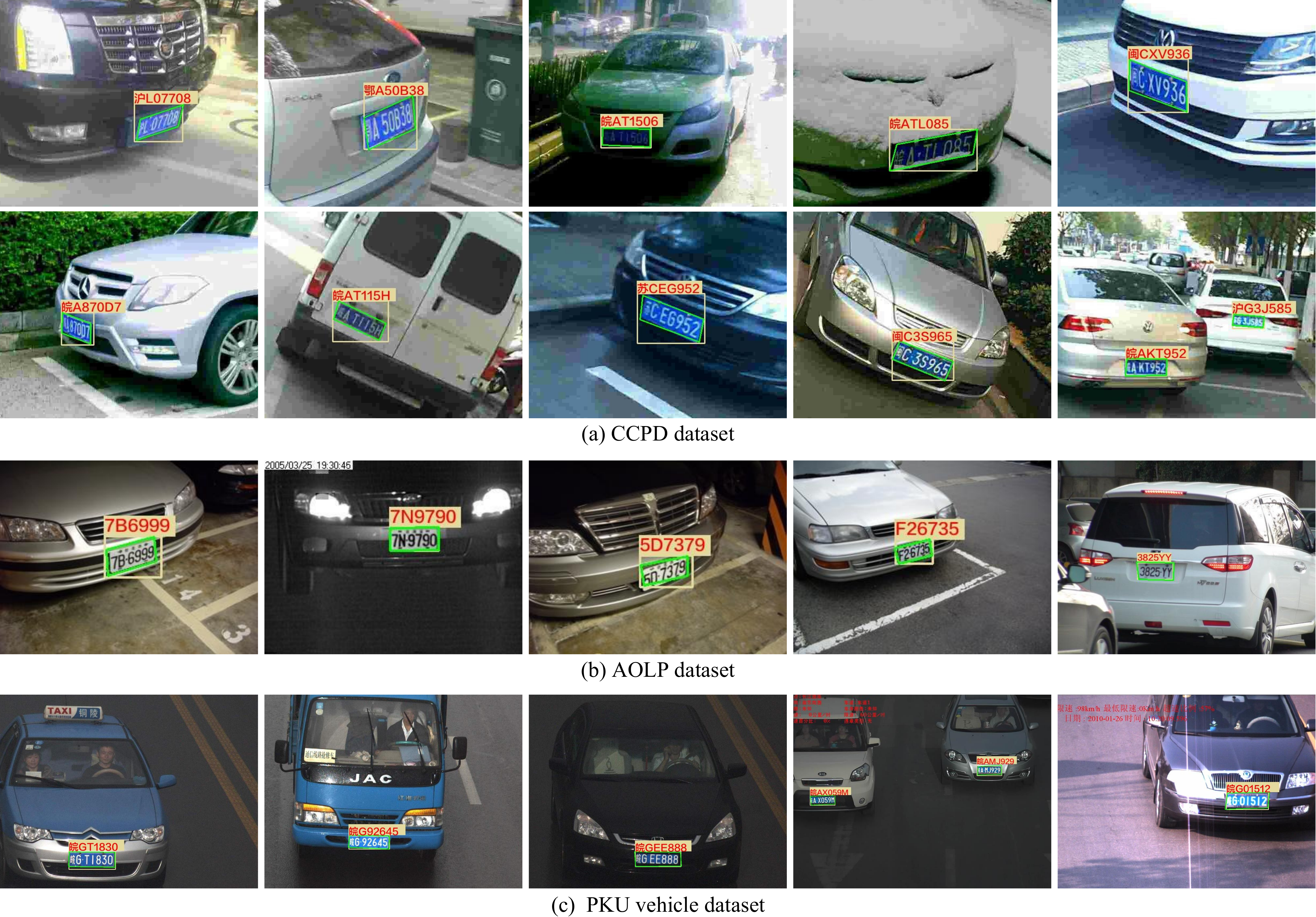}}
	\caption{Location and recognition results on three datasets. Images are cropped to small sizes for a better view. It should be noted that the results for PKU vehicle dataset are obtained from the model trained on CCPD dataset.
		\label{fig6}}
\end{figure*}

We compare our method with three other methods on this dataset, including Li \cite{Li2013}, Yuan \cite{Yuan2017}, TE2E \cite{Li2019} and WPOD\cite{Silva_2018_ECCV}. As shown in Table~\ref{tab4}, our method are more accurate and efficient than traditional detection methods provided by \cite{Li2013} and \cite{Yuan2017}. Compared with the deep learning method TE2E \cite{Li2019}, we achieve almost the same average performance. while use just $21 ms$ for inference and increase the speed to $47.6 fps$. Average performance of our method is slightly higher than that of WPOD \cite{Silva_2018_ECCV}. Summarily, our method is efficient for real-time applications. Example results are shown in Fig. \ref{fig6} (c).

\subsection{Performances on mixed dataset}

The generalization capabilities are always important for a convolutional neural network. We simply mix the training parts of CCPD (new version) and AOLP and train a unified model for testing. Then, we validate the model on these testing parts. The experimental results are shown in Table~\ref{tab5}. Baseline means results are from the models training with single dataset. It is easy to find that the unified model can be compatible with different kinds of license plates. The gap of AOLP is slightly bigger than that of CCPD due to sample imbalance. We also test PKU dataset which is not employed for training. Since both CCPD and PKU datasets have license plates from mainland China, the unified model can be directly employed on PKU dataset for detection theoretically. However, the performance is not as good as the baseline. Most of failed cases are license plates with yellow background which do not appear in the mixed training dataset. In addition, color jitterbugging for data augmentation is designed to improve the adaptability to different light environment. The range of color adjustment is relatively small. So, yellow plates cannot be generated from white or blue plates using the proposed color jitterbugging method. Summarily, we argue that a unified model can be compatible with different license plate layouts after training with diverse datasets. The diversity and distribution of training dataset can affect performance.

\begin{table}[!t]
	\renewcommand{\arraystretch}{}
	\caption{Performances (as percentages) on mixed dataset. Results for CCPD and AOLP are recognition accuracy in percentage. For PKU, results represent detection performances.}
	\label{tab5}
	\centering
	\begin{tabular}{lccc}
		\toprule	
		Dataset &Unified model(\%) &Baseline(\%) &Gap\\
		\midrule
		CCPD &96.91 &96.94 &0.03\\
		AOLP &98.40 &98.82 &0.42\\
		PKU &92.17 &99.80 &7.63\\
		\bottomrule
	\end{tabular}
\end{table}

\subsection{Ablation study}
The experiments of ablation study are totally based on new version of CCPD dataset (Base, Rotate, Tilt, Challenge), since the dataset is large and diverse enough to avoid occasionality. We make several groups of comparative experiments to evaluate the rationality and validity of the proposed method. Experimental results are shown in Table~\ref{tab6}. For all the experiments of ablation studies, ResNet-18 is adopted. 

\begin{table}[!t]
	\renewcommand{\arraystretch}{}
	\caption{Ablation study of the proposed method. We make comparisons of recognition accuracy between the model with and without rectification, FPN and post-processing. We make a comparison of detection accuracy between the model with and without recognition branch.}
	\label{tab6}
	\centering
	\resizebox{0.48\textwidth}{!}{
	\begin{tabular}{lcccc}
		\toprule	
		Performance(\%) &Base &Rotate &Tilt &Challenge\\	
		\midrule
		without rectification   &98.1 &87.2 &88.4 &84.6 \\
		with rectification &\textbf{99.2} &\textbf{97.9} &\textbf{95.5} &\textbf{92.9}\\
		\midrule
		without FPN  &98.6 &97.1 &94.3 &91.5 \\
		with FPN &\textbf{99.2} &\textbf{97.9} &\textbf{95.5} &\textbf{92.9}\\
		\midrule
		without Post-processing  &99.2 &97.3 &95.1 &91.7 \\
		with Post-processing &\textbf{99.3} &\textbf{97.9} &\textbf{95.5} &\textbf{92.9}\\
		\midrule
		without recognition  &99.2 &98.3 &95.4 &93.6 \\
		with recognition &\textbf{99.5} &\textbf{98.5} &\textbf{95.9} &\textbf{94.1}\\
		\bottomrule
	\end{tabular}}
\end{table}

To prove the validity of rectification with corners, we make a comparison between the model with and without rectification (Corners are regressed but not used). The experimental results demonstrate that the model with rectification can make more than $7\%$ improvements on Rotate, Tilt and Challenge sub-datasets with little extra calculation burden. To verify the effectiveness of FPN, we also make a comparison between the model with and without FPN. It is easy to find that the FPN module can increase performance by about $0.7\%\sim 1.2\%$. 
To prove the effectiveness of post-processing, a comparison is made between the performances before and after post-processing. Results indicate that the accuracy can be improved by post-processing. Generally, the more difficult the sample is, the more obvious the improvement is. In addition, we compare the detection performance between the model with and without recognition branch. We find that the end-to-end method can improve the detection task to a certain extent by using multi-task training and shared features. 

We also test the influence of the backbone. In Table~\ref{tab2}, we present the performances for two backbones, ResNet-18 and ResNet-50. The heaver backbone can just bring $0.4\%$ improvement for average performance, while significantly reduces the speed from $36fps$ to $26fps$. 

\subsection{Speed Analysis}

\begin{table}[!t]
	\renewcommand{\arraystretch}{}
	\caption{Time consumption analysis (in $ms$) based on CCPD dataset. The total time is divided into three parts: feature extraction, location and recognition.}
	\label{tab7}
	\centering
	\resizebox{0.48\textwidth}{!}{
	\begin{tabular}{lcccc}
		\toprule	
		Method &Feature ext. &Location &Recognition &Total\\
		\midrule
		Ours (ResNet-18) &12.2 &8.6 &7.1 &  27.9\\
		Ours (ResNet-50) &22.6 &8.9 &7.2 &  38.7\\
		\bottomrule
	\end{tabular}}
\end{table}

\begin{table}[!t]
	\renewcommand{\arraystretch}{}
	\caption{Speed analysis on different hardware platforms.}
	\label{tab8}
	\centering
	\resizebox{0.48\textwidth}{!}{
		\begin{tabular}{lccc}
			\toprule	
			Target platform &Size ($w\times h$) &Accuracy (\%) &FPS \\
			\midrule
			CPU1 &$640\times 1024$ &99.3 &3.4\\
			CPU2 &$640\times 1024$ &99.3 &7.3\\
			GPU1 &$640\times 1024$ &99.3 &21.3\\
			GPU2 &$640\times 1024$ &99.3 &35.7\\
			Jetson Nano &$640\times 1024$ &99.3 &0.62\\
			Jetson Nano (with TRT) &$640\times 1024$ &99.3 &2.8\\
			Jetson Nano &$320\times 512$ &98.6 &1.97\\
			Jetson Nano (with TRT) &$320\times 512$ &98.6 &8.1\\
			\bottomrule
	\end{tabular}}
\end{table}

Firstly, we make a time consumption analysis of the proposed framework in different stages based on CCPD dataset (new version). The resolution of images is set to $640\times 1024$. As shown in Table~\ref{tab7}, we simply divide the total time into three parts (feature extraction, location and recognition) that are executed sequentially. It should be noted that time costs of RoIAlign and rectification are included in recognition. Post-processing is also merged into recognition since the time cost is only about $0.1$ ms. The time costs of location and recognition are almost not changed with different backbones. Compared with most of the two-step solutions and RNN-based methods, our method is more efficient in both location and recognition.

Then, we analyze the speed of the proposed method on different hardware platforms, including CPU1 (Intel Core(TM) i7 7700, 3.60GHZ), CPU2 (Intel(R) Xeon(R) W-2133, 3.60GHz), GPU1 (Nvidia GTX 1060, 6GB), GPU2 (Nvidia RTX 2080Ti, 11GB) and NVIDIA Jetson Nano (a low-power embedded platform for edge computing). The target platforms can be generally divided into three categories: desktop computer (CPU1, GPU1), computing server (CPU2, GPU2) and edge device. ResNet-18 is used for experiments. As shown in Table~\ref{tab8}, our method can run in real-time on GPU platforms. However, it is much slower on CPU platforms, due to the computational complexity of deep neural networks.  Jetson Nano (with TRT) is a Tensor-RT version of our method to speed up the computation. With TRT, our method can run on the Jetson Nano at about 2.8 fps. However, when we reduce the input image to half size ($320\times 512$), our method can run at about 8 fps. Meanwhile, the accuracy drops from $99.3\%$ to $98.6\%$, which is generally acceptable. As we know, Jetson Nano is an entry-level computing platform with only 472 GFlops computational power. Most of the smart cameras and edge devices are more powerful. Accordingly, we argue that our method is also suitable for smart cameras and edge devices.

\section{Conclusion}
\label{conclusion}
In this work, we have proposed an efficient and accuracy framework to locate and recognize car license plates using a single neural network. For location, both bounding boxes and corners are regressed by an anchor-free detection head. For recognition, RoIAlign and rectification are used to extract the accurate feature areas of license plates. Then, a small network is designed to decode sequences of characters directly by beam search. It is a concise and end-to-end framework that can directly recognize license plates in unconstrained scenes without car detection, plate and character segmentation. In addition, it is efficient enough for real-time and edge computing applications. Comprehensive experiments on several public datasets for detection and recognition demonstrate the advantages of our method. One deficiency of this approach is that it requires large amounts of training data with corner-level annotations. For future work, we plan to solve multi-line license plate recognition problems and migrate this method to other applications, such as text spotting.

\section{Conflict of Interest}
The authors declare that they have no conflict of interest to this work.


%
%


\end{document}